# A NEW APPROACH FOR MEASURING SEMANTIC SIMILARITY OF ONTOLOGY CONCEPTS USING DYNAMIC PROGRAMMING

[1]ABDELHADI DAOUI, [2]NOREDDINE GHERABI AND [3]ABDERRAHIM MARZOUK

[13] Hassan 1st University, FSTS, Department of Mathematics and Computer Science,
IR2M Laboratory, Settat, Morocco

[2] Hassan 1st University, ENSAK, Department of Computer Science,
LIPOSI Laboratory, Khouribga, Morocco

E-mail: [1]abdo.daoui@gmail.com, [2]gherabi@gmail.com, [3]amarzouk2004@yahoo.fr

## ABSTRACT

Today, with the emergence of semantic web technologies and increasing of information quantity, searching for information based on the semantic web has become a fertile area of research. For this reason, a large number of studies are performed based on the measure of semantic similarity. Therefore, in this paper, we propose a new method of semantic similarity measuring which uses the dynamic programming to compute the semantic distance between any two concepts defined in the same hierarchy of ontology. Then, we base on this result to compute the semantic similarity. Finally, we present an experimental comparison between our method and other methods of similarity measuring.

## 1. INTRODUCTION

Recently, with explosion of information quantity in the web and the richness of natural languages used during the redaction of this information, the traditional search based on keywords has become useless (does not meet the needs of Internet users because the result depends on the keywords chosen by Internet users themselves). Therefore, a new type of search which is based on the semantic web [1] has become a necessity. Thus, the search engines that want to implement this new type of search should be able to measure the semantic similarity. For this reason, the semantic similarity measuring has become a fertile domain of research.

In this paper, our work focuses on semantic similarity measuring between concepts of ontology. This concept is a base of several works of research. The authors of [2] propose a method to compute the semantic similarity between words using a multiple information resources (lexical, corpus and taxonomy). Jeffrey Hau, William Lee and John Darlington in [3] present a method to define the compatibility between semantic web services [4] [5] which are annotated by OWL ontologies (Web Ontology language [6]) using the semantic similarity. Also, we can find this type of measure between ontologies [7] [8], for example the authors of [8] propose a new method that allows computing the semantic similarity between two ontologies in three steps. In the first step, the authors compute the semantic similarity between the nodes of the two ontologies. Then, they compute the semantic similarity between the relations of the two ontologies. At last, the authors combine these two previous results to form one unified value which represents the semantic similarity computed between these two ontologies.

Our proposed method aims to compute the semantic similarity between any two concepts in the same hierarchy of ontology (in this paper we use the term graph to describe ontology). For this, in the first step, we generate a routing table which contains all possible paths from these nodes to the root node. Then, we define the shortest paths on which we will apply the dynamic programming to obtain the value of semantic similarity between these two nodes. In [9] the authors base on the same technique (dynamic programming) for computing the rate of similarity between outlines of 2D shapes using the extraction of XML data which represent local and global features of these outlines. Also,

Pelin Dogan, Markus Gross and Jean-Charles Bazin in [10] have used the dynamic programming technique for temporal aligning of video frames with narrative sentences (the descriptions of natural language accompanying these videos). To do that, the authors relied on textual and visual information that provides automatic timestamps for each narrative sentence.

The rest of the current paper is organized as follows: section 2 presents our method. Then, section 3 provides the experimental comparison with some other methods of similarity measuring. Finally, section 4 is devoted to our conclusion.

## 2. PROPOSED METHOD

For computing the semantic similarity between two concepts present in the same hierarchy of ontology, we have designed a new method which is summarized in figure 1. This method supports the ontologies which use only the relations of type "is-a" (inheritance). Therefore, our graph will be an oriented graph towards the root node.

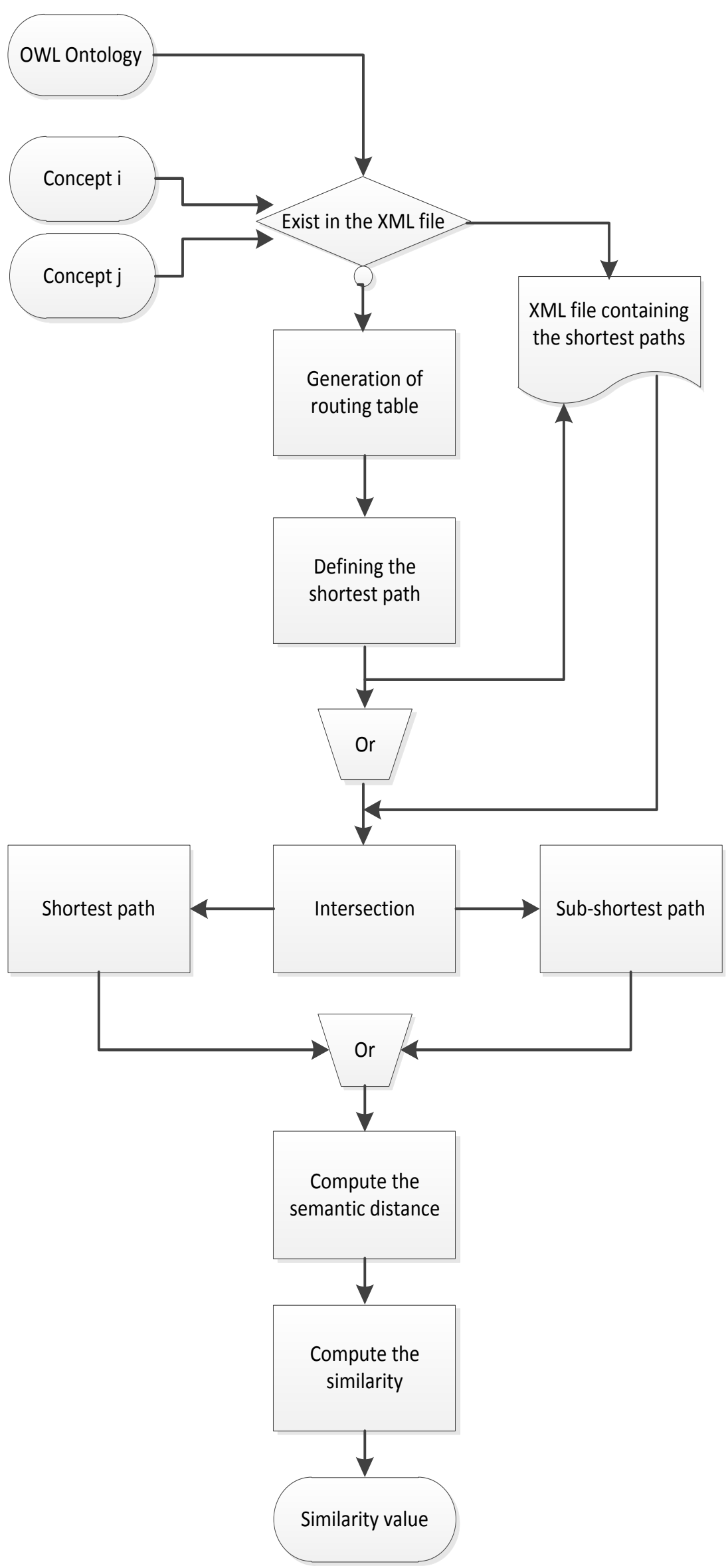

*Figure 1: A Graphical Representation of the Proposed Algorithm.*

Our designed method allows, in the first place, the generation of the routing table for the two concepts which we need to compute the semantic similarity between them and defining the shortest paths (section 2.1). Then, we use the dynamic programming technique to calculate the semantic distance between these two concepts (section 2.2). Finally, in section 2.3 we compute the semantic similarity between using the semantic distance calculated in the previous section.

## 2.1 Routing Table Generation

The routing table represents a table containing the two nodes that we need to measure the semantic similarity between them and all possible paths from these nodes to the root node.
We consider the following ontology:

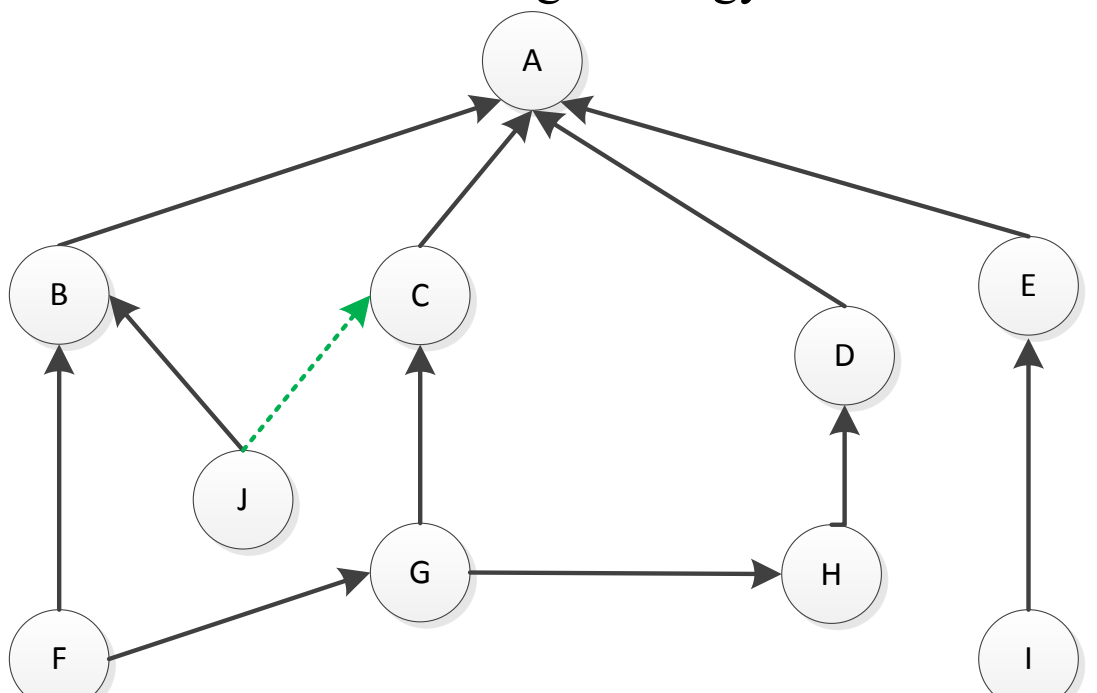

*Figure 2: An Example of Ontology.*

For example, we need to generate the routing table for the node F presented in figure 2.

*Table 1: Routing Table*

| Nodes | All paths to the root node |
|---|---|
| F | (F,B,A);(F,G,C,A);(F,G,H,D,A) |

By analysis of this routing table, we have found three paths ((F, B, A);(F, G, C, A);(F, G, H, D, A)) between the node F and the root node. The shortest path will be the path, which has the minimum number of nodes among the generated paths in the routing table. In our example the shortest path between the node F and the root node is (F, B, A).

If we have two equal paths between the node J and the root node (figure 2), the shortest path for the node J, which will be used, is the one that has more nodes in common with the second concept with which we want to compute the semantic similarity. For example, the shortest path for the node F is (F, B, A) and for the node J, we have two equal paths (J, B, A) and (J, C, A). But we will choose (J, B, A) because this path has more nodes in common with the concept (F, B, A) compared to (J, C, A). This technique allows minimizing the semantic distance between the concepts that we want to compute the semantic similarity between them.

To avoid redefining the shortest path for a given node more than once, we proposed to use an XML file to store the shortest paths computed for potential use in the future. This XML file follows the structure of figure 3.

```
<?xml version="1.0" encoding="utf-8" ?>
<SPaths> //the list of calculed shortest paths
    <SPath ID="F">  //a shortest path from a given
node to the root node
        <SNode> </SNode>  //the start node
        <Node> </Node>
        <Node> </Node>
         .
         .
         .
        <RNode> </RNode>  //the root node
    </SPath>
    .
    .
    .
</SPaths>
```

*Figure 3: XML File for storing the shortest paths*

For validating this XML file, we use the following DTD file:

```
<!ELEMENT SPaths (SPath)>
<!ELEMENT SPath (RNode,Node*,SNode?)>
<!ELEMENT SNode (#PCDATA)>
<!ELEMENT Node (#PCDATA)>
<!ELEMENT RNode (#PCDATA)>
<!ATTLIST SPath id ID #REQUIRED>
```

*Figure 4: The corresponding DTD file.*

## 2.2 Semantic Distance Calculation

At this level, for computing the semantic distance between two concepts we use the dynamic programming technique to obtain the value of alignment between the two sequences of symbols composing the shortest path for each of these two concepts. The value of alignment will be the semantic distance between these two concepts. In this paper, we use an algorithm of dynamic programming called Levenshtein Edit Distance [14], this algorithm allows computing the distance between two strings through a set of operations, where an operation can be a substitution, a deletion or an insertion of a single character. But we are adapting it to support execution of these three operations on a motif (node designation) that can be a single character or even a word.

Our adapted algorithm for computing the sematic distance between two concepts is designed as follows:

$$D(\mathrm{i},\mathrm{j}) = \min \begin{Bmatrix} D(i-1, j-1) + subCost \quad // a\ substitution \\ D(\mathrm{i}-1,\mathrm{j}) + \mathrm{F} + \varepsilon(\mathrm{i}-1) \quad // \text{ a deletion} \\ \mathrm{D}(\mathrm{i},\mathrm{j}-1) + \mathrm{F} + \varepsilon(\mathrm{j}-1) \quad // \text{ an insertion} \end{Bmatrix} \quad (1)$$

D (i, j) represents the value of edit distance in position (i, j), F represents a fixed penalty with a value equal to one and subCost represents a cost. This cost is equal to zero if the nodes that exist in the positions (i-1) and (j-1) are identical and equal to 1.5+ ε(i-1)+ ε(j-1) if they are not identical.

$$\varepsilon(\mathrm{n}) = [\mathrm{depth}(\mathrm{n}) + \frac{N(n)}{NTNodes(\mathrm{G}) + 1} + 1]^{-1} \qquad (2)$$

Where, ε(n) represents the weight of the node n, depth (n) represents the depth of the node n in the shortest path (the depth of the root node is equal to zero), N(n) represents the order number of this node between their siblings in the graph G (this number begins from zero). And NTNodes(G) represents the total number of nodes in the graph G.

The formula 1 will be used for computing the matrix M[0,…,m;0,…,n], where m and n represent the length of sequences to be compared. We find the semantic distance value in M [m; n].

```
Function semanticDistance(sequence1,sequence2)
// sequence represents the nodes designations
// list that constitute the SPath
For i=0 to m do
    M(i,0) = i
End for
For j=0 to n do
    M(0,j) = j
End for
For i=1 to m do
    For j=1 to n do
        if sequence1 (i-1)== sequence2 (j-1) then
                subCost=0
            else
                subCost=1.5+ ε(i-1)+ ε(j-1)
            End if
            delCost=1+ ε(i-1)
            insertCost=1+ ε(j-1)
```

$$\mathrm{M}(\mathrm{i},\mathrm{j}) = \min \begin{Bmatrix} D(i-1, j-1) + subCost \\ D(\mathrm{i}-1,\mathrm{j}) + delCost \\ \mathrm{D}(\mathrm{i},\mathrm{j}-1) + insertCost \end{Bmatrix}$$

```
    End for
End for
Return M(m,n) //the semantic distance value
```

*Figure 5: Our Adapted Algorithm for Semantic Distance Computing.*

Consider the ontology presented in figure 2, if we want to compute the semantic distance between the nodes I and F, we need to define the shortest path between these nodes and the root node (in our case is A) by using the shortest path function. After execution, we obtain (A, E, I) and (A, B, F) like a result. Then, we compare these two sequences by using the dynamic programming:

*Table 2: An Example of Application of Dynamic Programming Algorithm.*

| | | A | E | I |
|---|---|---|---|---|
| | 0 | 1 | 2 | 3 |
| A | 1 | 0 | 1.44 | 2.773 |
| B | 2 | 1.5 | 2.44 | 3.773 |
| F | 3 | 2.833 | 3.773 | 4.606 |

After the computation of matrix M, we can see that the semantic distance between the concepts (A, E, I) and (A, B, F) is equal to 4.606.

For minimizing the execution time of semantic distance calculation, we have designed an intermediate phase, which execute before semantic distance calculation and allows deleting the all common nodes between SPath1 and SPath2 except one node that verifies these two conditions:

1. All nodes either in SubSPath1 or SubSPath2 are connected.
2. SubSPath1 ∩ SubSPath2 = n.

Where, $SPath_i$ represents the shortest path between a given node and the root node, $SubSPath_i$ represents the sub shortest path of $SPath_i$ that we find after deleting the all common nodes from SPath1 and SPath2 except the node n, which must remain in common between these two sub shortest paths.

For example, we consider the graph shown previously in figure 2. In this graph, we have the two shortest paths for the nodes J and F are (J, B, A) and (F, B, A). The computation of semantic distance on these paths is burdensome either for execution time or for memory because the calculation applied to node A is useless (especially if we have a long set of nodes in common). For this, we can use the sub shortest paths (J, B) and (F, B) in their place.

### 2.3 Semantic Similarity Calculation

In this section, we exploit the semantic distance computed previously in this paper to define the semantic similarity using an inverse relation between these two concepts (semantic distance and semantic similarity). By analyzing the output value of the semantic similarity function, we can categorize it in three categories:

1. The two concepts are the same.
2. Nothing in common between them.
3. There is a rate of semantic similarity between them.

Therefore, this function should verify three conditions:

1. $\forall (A,B) \in G : 0 \le SSim(\mathrm{A,B}) \le 1$
2. $\forall A \in G : SSim(\mathrm{A,A}) = 1$
3. $\forall (A,B,C) \in G : if\ SDis(\mathrm{A,B}) > \mathrm{SDis(A,C)}$ then $\mathrm{SSim(A,B)} < \mathrm{SSim(A,C)}$

Where A, B and C represent three concepts of graph G, SSim represents the semantic similarity and SDis represents the semantic distance.
To compute the semantic similarity in the current paper, we have used the function proposed in [13].

$$SSim(A,B) = \frac{1}{\deg * SDis(A,B) + 1} \qquad (0 < \deg \le 1) \qquad (3)$$

A and B represent two concepts that we want to compute the semantic similarity between them, the parameter "deg" represents the impact degree of Semantic distance on semantic similarity and its concrete value will be defined in the experience phase.

### 2.4 Global Algorithm

Global algorithm represents the algorithm of our proposed method for computing the semantic similarity between two concepts of hierarchical ontology (all their concepts are connected by the same relation type "is-a").

```
Input: C1, C2, deg
Output: semantic similarity value

If C1==C2 then
    SDis=0
Else
    Generate the routing table for C1 and C2
    Define SPath1 and SPath2
    SDis= semanticDistance(SPath1, SPath2)
End if
```

$$SSim = \frac{1}{\deg * SDis + 1}$$

```
Return SSim
```

*Figure 6: Our Proposed Algorithm.*

## 3. EXPERIMENTS

This section is devoted to experimental comparison between our proposed method and two other methods of semantic similarity measuring. For this, we have used a fragment of ontology hierarchy shown in figure 7.

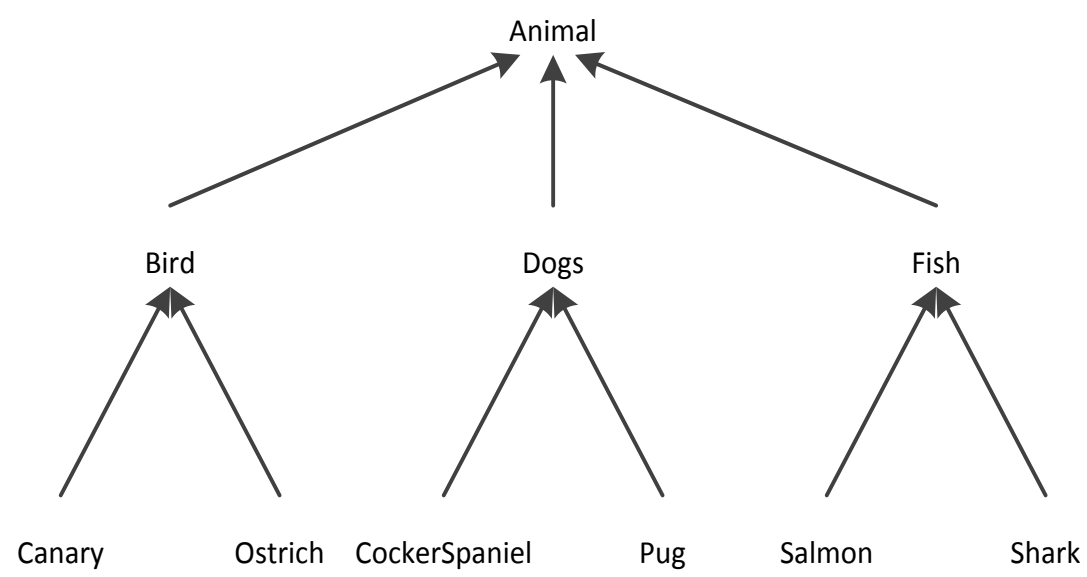

*Figure 7: A Fragment of Ontology Hierarchy.*

For a better interpretation of our method we applied the algorithm of similarity computation on the example of figure 7.

The tables below give an overview of the calculation of semantic similarity by applying our method and comparing with other methods where we set the parameter "deg" to 0.2. Each concept is defined by a starting node and an arrival node (last node), in the tables the concepts are represented by the last node. For example, the concept (Animal, Fish, Shark) will be represented by the node called Shark.

*Table 3: Computing Semantic Similarity Using [15].*

| | Animal | Bird | Fish | Shark | Canary | Ostrich | Pug |
|---|---|---|---|---|---|---|---|
| Animal | 1 | 0 | 0 | 0 | 0 | 0 | 0 |
| Bird | 0 | 1 | 0 | 0 | 0 | 0 | 0 |
| Fish | 0 | 0 | 1 | 0 | 0 | 0 | 0 |
| Shark | 0 | 0 | 0 | 1 | 0 | 0 | 0 |
| Canary | 0 | 0 | 0 | 0 | 1 | 0 | 0 |
| Ostrich | 0 | 0 | 0 | 0 | 0 | 1 | 0 |
| Pug | 0 | 0 | 0 | 0 | 0 | 0 | 1 |

*Table 4: Computing Semantic Similarity Using [13].*

| | Animal | Bird | Fish | Shark | Canary | Ostrich | Pug |
|---|---|---|---|---|---|---|---|
| Animal | 1 | 0,71 | 0,71 | 0,58 | 0,58 | 0,58 | 0,58 |
| Bird | 0,71 | 1 | 0,55 | 0,47 | 0,76 | 0,76 | 0,47 |
| Fish | 0,71 | 0,55 | 1 | 0,76 | 0,47 | 0,47 | 0,47 |
| Shark | 0,58 | 0,47 | 0,76 | 1 | 0,41 | 0,41 | 0,41 |
| Canary | 0,58 | 0,76 | 0,47 | 0,41 | 1 | 0,62 | 0,41 |
| Ostrich | 0,58 | 0,76 | 0,47 | 0,41 | 0,62 | 1 | 0,41 |
| Pug | 0,58 | 0,47 | 0,47 | 0,41 | 0,41 | 0,41 | 1 |

Our method is based on two steps to compute the semantic similarity.
These steps are defined as follows:
Step 1: computing the semantic distance.

*Table 3: Semantic Distance Computing Using Dynamic Programming.*

| | | Animal | Fish | Shark |
|---|---|---|---|---|
| | 0 | 1 | 2 | 3 |
| Animal | 1 | 0 | 1.458 | 2.782 |
| Bird | 2 | 1.5 | 2.458 | 3.782 |

Step 2: computing the semantic similarity.

The formula for calculating semantic similarity is defined as follows:

SSim (C1, C2)=1/(0.2*3.782+1) = 0.569

We follow these two steps to compute the semantic similarity using our method between each two ontological concepts and the result of similarity is stored in Table 6.

*Table 4: Computing Semantic Similarity Using Our Method.*

| | Animal | Bird | Fish | Shark | Canary | Ostrich | Pug |
|---|---|---|---|---|---|---|---|
| Animal | 1 | 0.769 | 0.774 | 0.643 | 0.638 | 0.639 | 0.641 |
| Bird | 0.769 | 1 | 0.67 | 0.569 | 0.79 | 0.791 | 0.568 |
| Fish | 0.774 | 0.67 | 1 | 0.791 | 0.569 | 0.569 | 0.571 |
| Shark | 0.643 | 0.569 | 0.791 | 1 | 0.52 | 0.521 | 0.522 |
| Canary | 0.638 | 0.79 | 0.569 | 0.52 | 1 | 0.699 | 0.519 |
| Ostrich | 0.639 | 0.791 | 0.569 | 0.521 | 0.699 | 1 | 0.519 |
| Pug | 0.641 | 0.568 | 0.571 | 0.522 | 0.519 | 0.519 | 1 |

Table 3 represents the results of the first method [15] which can only find the similarity between the same concepts, table 4 represents the results of the second method [13] which can find a rate of similarity between concepts of ontology as our proposed method, where our results are presented in the table 6. In contrast to our method, the method presented in table 4, gives the same rate of semantic similarity between a given concept and all the concepts, which represent a direct specification of this concept. For example the semantic similarity between the concept (Animal) and the concepts (Animal, Bird) and (Animal, Fish) is equal to 0.71. But in the reality, the semantic similarity between these concepts is different. For this reason, the rates of semantic similarity computed using our proposed method on the same concepts are different. For example, the rate of semantic similarity between (Animal) and (Animal, Bird) is equal to 0,769 and between (Animal) and (Animal, Fish) is equal to 0.774.

Also, the semantic similarity between two concepts more specific is greater than two others more generalist. For example, the semantic similarity between the concepts (Animal, Bird, Canary) and (Animal, Bird, Ostrich) is greater than the semantic similarity between (Animal, Bird) and (Animal, Fish).

In addition, we use the XML file for storing the shortest paths already computed. This technique allows minimizing the execution time that can be reserved every time for redefining these shortest paths.

The method proposed in this paper is based on a dynamic programming algorithm, which is well known by its robustness and speed. Therefore, this one can compute the semantic similarity between concepts of hierarchical ontology in very short time even in industrial size ontologies.

## 4. CONCLUSION

In this paper, we have presented a new method for computing the semantic similarity between two concepts of the same ontology using the dynamic programming. Then, we have compared it against other methods of semantic similarity measuring and the obtained results are very interesting.

In the future works, we are interested to compute the semantic similarity between concepts in different ontologies.

## REFERENCES

[1] T. B. Lee, J. Hendler, O. Lassila, "The semantic web", *Scientific America*, 2001, pp. 1-18.

[2] Y. Li, Z. A. Bandar, D. McLean, "An Approach for Measuring Semantic Similarity between Words Using Multiple Information Sources", *IEEE Transactions on Knowledge and Data Engineering*, Vol. 15, No. 4, 2003, pp. 871-882.

[3] J. Hau, W. Lee, J. Darlington, "A Semantic Similarity Measure for Semantic Web

Services", *the 14th international conference on World Wide Web*, Chiba, Japan, 2005.

[4] M. Burstein, C. Bussler, M. Zaremba, T. Finin, M. N. Huhns, M. Paolucci, A. P. Sheth, S. Williams, "A Semantic Web Services Architecture", *IEEE Internet Computing*, Vol. 9, No. 5, 2005, pp. 52-61.

[5] S. A. McIlraith, D. L. Martin, "Bringing Semantics to Web Services", *IEEE Intelligent Systems*, Vol. 18, No. 1, 2003, pp. 90-93.

[6] https://www.w3.org/TR/owl2-overview, last accessed 17/11/2016.

[7] F. Li, "An Improved Method about the Similarity Calculation of Ontology", *International Conference on Multimedia Technology, IEEE*, 2010, pp. 1-4.

[8] H. Wang, X. Han, "Research on Similarity of Semantic Web", *International Conference on Computer Application and System Modeling, IEEE,* 2010*,* pp. 166-169.

[9] N. Gherabi, M. Bahaj, "Outline Matching of the 2D Shapes Using Extracting XML Data", *ICISP 2012, Springer, Heidelberg*, 2012, pp. 502–512.

[10] P. Dogan, M. Gross, J.C. Bazin, "Label-Based Automatic Alignment of Video with Narrative Sentences". *ECCV 2016, Springer International, Switzerland,* 2016, pp. 605–620.

[11] J. Zhong, H. Zhu, J. Li, Y. Yu, "Conceptual Graph Matching for Semantic Search", *ICCS 2002, Springer, Heidelberg*, 2002, pp. 92-106.

[12] Y. Ganjisaffar, H. Abolhassani, M. Neshati, M. Jamali, "A Similarity Measure for OWL-S Annotated Web Services", *International Conference on Web Intelligence, IEEE*, 2006, *pp.* 621-624.

[13] J. Ge, Y. Qiu, "Concept Similarity Matching Based on Semantic Distance", the *4th International Conference on Semantics, Knowledge and Grid, IEEE,* 2008, pp. 380-383.

[14] E. S. Ristad, P. N. Yianilos , " Learning String-Edit Distance", *IEEE Transactions on Pattern Analysis and Machine Intelligence,* Vol. 20, No. 5, 1998,522-532.

[15] F. Giunchiglia, P. Shvaiko, M. Yatskevich , "S-Match: an algorithm and an implementation of semantic matching", *ESWS 2004, Springer, Heidelberg,* 2004, pp. 61-75.